\pdfoutput=1

\documentclass[11pt]{article}

\usepackage[final]{acl}

\usepackage{times}
\usepackage{latexsym}
\usepackage{amsmath,amsfonts}
\usepackage{algorithmic}
\usepackage{algorithm}
\usepackage{array}
\usepackage{textcomp}
\usepackage{stfloats}
\usepackage{url}
\usepackage{verbatim}
\usepackage{graphicx}
\usepackage{threeparttable}
\usepackage{booktabs}
\usepackage{multirow}
\usepackage{xcolor,colortbl}
\usepackage[T1]{fontenc}

\usepackage[utf8]{inputenc}

\usepackage{microtype}

\usepackage{inconsolata}

\title{Key-Point-Driven Mathematical Reasoning Distillation of Large Language Model}

\author{Xunyu Zhu$^{1,2}$, Jian Li$^{3}$\thanks{Corresponding author}, Can Ma$^{1,2}$, Weiping Wang$^{1,2}$\\
$^1$Institute of Information Engineering, Chinese Academy of Sciences\\
$^2$School of Cyber Security, University of Chinese Academy of Sciences\\
$^3$ School of Artificial Intelligence, Beijing Normal University\\
\texttt{\{zhuxunyu, macan, wangweiping\}@iie.ac.cn, jli@bnu.edu.cn}\\
}

\begin{document}
\maketitle
\begin{abstract}
Large Language Models (LLMs) excel in mathematical reasoning due to their extensive parameters and training data, but their high computational demands hinder deployment. Distilling LLM reasoning into Smaller Language Models (SLMs, $\le$ 1B parameters) is a potential solution, yet these models often struggle with calculation and semantic errors. Previous work introduced Program-of-Thought Distillation (PoTD) to reduce calculation mistakes. To tackle semantic errors, we propose Key-Point-Driven Mathematical Reasoning Distillation (KPDD), which improves SLM reasoning by splitting the problem-solving process into Key Points Extraction and Step-by-Step Solution. KPDD includes KPDD-CoT, generating Chain-of-Thought rationales, and KPDD-PoT, producing Program-of-Thought rationales. Experiments show KPDD-CoT enhances reasoning capabilities, while KPDD-PoT achieves state-of-the-art performance in mathematical tasks, effectively reducing misunderstanding errors and promoting the deployment of efficient, capable SLMs.
\end{abstract}

\section{Introduction}
Large language models (LLMs)~\cite{abs-2303-08774,yang2024qwen2technicalreport,abs-2401-02954,abs-2401-04088,abs-2406-12793,abs-2403-17297} have achieved impressive performance in mathematical reasoning tasks. Recent work~\cite{0002WSLCNCZ23} further proposes Chain-of-Thought (CoT) to enhance the mathematical reasoning abilities of LLMs. However, the massive scale of LLMs presents significant challenges for deployment.

\begin{figure}[!ht]
	\centering
	\includegraphics[width=\columnwidth]{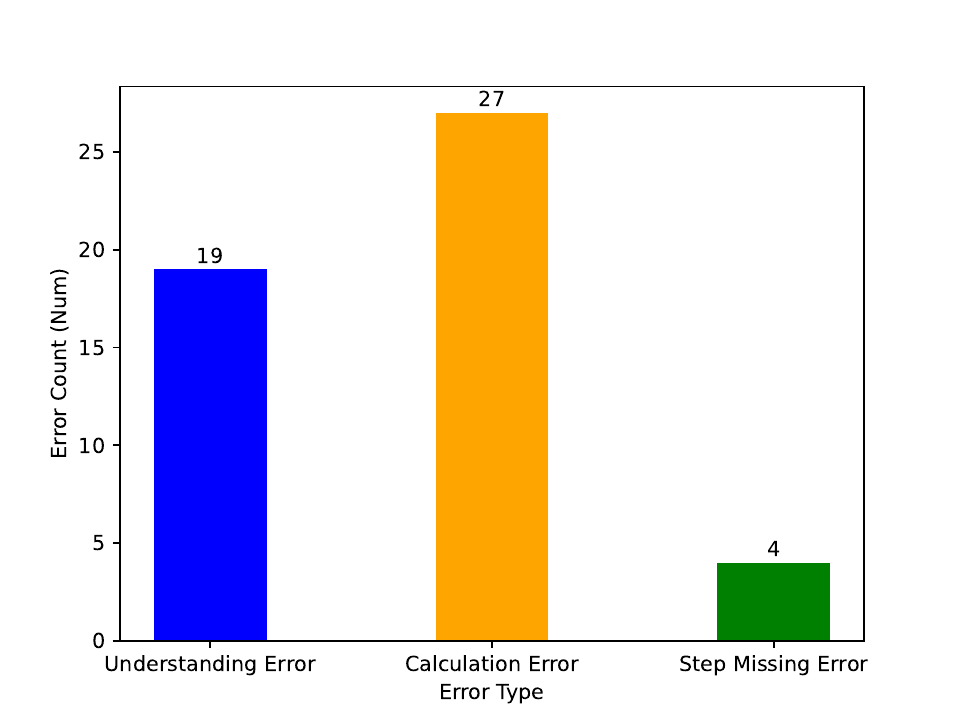}
	\caption{\textbf{Error analysis of 50 GSM8K problems with incorrect answers returned by CoTD using FlanT5-Base.} The experimental results indicate that understanding errors and calculation errors are the major factors affecting CoTD's reasoning performance.}
	\label{fig:cotd-error-analysis}
\end{figure}

A feasible solution to address this problem is to use black-box distillation to transfer mathematical reasoning abilities from LLMs to SLMs. Chain-of-Thought Distillation (CoTD)~\cite{ho-etal-2023-large} is a representative mathematical reasoning distillation method. CoTD prompts LLMs to generate reasoning processes for each question, constructing a mathematical reasoning dataset. This dataset is then used to fine-tune SLMs, enhancing their mathematical reasoning abilities. However, there remains a significant performance gap between SLMs and LLMs. Prior work~\cite{wei2022chain} identifies three main error types in CoT reasoning: Calculation errors, Missing Step errors, and Semantic misunderstanding errors. To explore the reasons for the performance gap between SLMs and LLMs, we conducted the same error analysis on CoTD. Our preliminary experiments (shown in Figure~\ref{fig:cotd-error-analysis}) indicate that the most common issues affecting CoTD's mathematical reasoning performance are calculation errors and semantic misunderstandings. Prior work~\cite{abs-2305-13888} proposed Program-of-Thought Distillation (PoTD) to avoid calculation errors by formulating the reasoning process as a Python program executed by an external interpreter. This approach allows the SLM to focus on generating the program, avoiding calculation errors and improving reasoning performance. Given these circumstances, our paper focuses on addressing semantic misunderstanding errors in CoTD to further enhance the reasoning performance of SLMs.

In our paper, we propose a novel mathematical reasoning distillation method called Key-Point-Driven Mathematical Reasoning Distillation (KPDD) to enhance the mathematical reasoning performance of SLMs. KPDD breaks the reasoning process into two parts: (1) Key points Extraction: identifies and extracts the core question and the relevant information from the original problem. (2) Step-by-Step Solution: uses the extracted key points to solve the problem in a step-by-step manner. The second part is further divided into two formats, KPDD-CoT and KPDD-PoT: (1) KPDD-CoT: Generates rationales in the form of Chain-of-Thought (CoT). This method focuses on reducing misunderstanding errors and explicitly illustrates the reasoning process, aiding in error analysis. (2) KPDD-PoT: Generates rationales in the form of Program-of-Thought (PoT). This approach not only reduces misunderstanding errors but also avoids calculation errors, further enhancing the SLM's mathematical reasoning performance.

We use KPDD to fine-tune FlanT5 models, and evaluate these models on several mathematical reasoning dataset, including GSM8K, ASDiv, SVAMP, and MultiArith. Our experiment results show that KPDD-CoT significantly enhances SLMs' reasoning abilities, while KPDD-PoT enables SLMs to achieve state-of-the-art (SOTA) mathematical reasoning performance.  Furthermore, our error analysis on KPDD confirms that KPDD effectively mitigates misunderstanding errors, thereby improving the mathematical reasoning performance of SLMs.

Our contributions are summarized as follows:
\begin{enumerate}
	\item Our study reveals that misunderstanding errors and calculation errors are the major factors limiting CoTD's reasoning.
	\item We propose Key-Point-Driven Mathematical Reasoning Distillation (KPDD) to alleviate misunderstanding errors and effectively improve the reasoning performance of SLMs.
	\item Extensive experiments show that KPDD outperforms other methods across various benchmarks and achieves new state-of-the-art results on these mathematical reasoning datasets.
\end{enumerate}

\section{Related Work}
\subsection{Chain-of-Thought Reasoning} 
Chain-of-Thought refers to prompt LLMs to solve the question step by step. Prior work~\cite{wei2022chain} finds that Chain-of-Thought can effectively improve the reasoning performance of LLMs. 
Based on the findings, ~\citet{NEURIPS2022_8bb0d291} further introduce zero-shot CoT, which significantly improves the model's reasoning performance by simply adding the prompt "Let's think step by step" before answering. 
To avoid calculation error in CoT, ~\citet{chen2023program} formulate the reasoning process into program. ~\citet{0002WSLCNCZ23} introduce a self-consistency decoding strategy, which generates diverse reasoning paths and then selects the most consistent answer by considering these paths comprehensively. Least-to-most prompting~\cite{zhou2023leasttomost} breaks down complex problems into a series of simpler subproblems and solves these subproblems sequentially. 
 ~\citet{abs-2404-14963} encourage LLMs to deeply understand problems and leverage useful information for better reasoning. 
Inspired by these methods, our work introduces Key-Point-Driven Mathematical Reasoning Distillation (KPDD), which generate distillation dataset with abundant mathematical reasoning knowledge to enhance SLMs' mathematical reasoning.

\subsection{Black-box Distillation}
Currently, LLMs usually have stronger reasoning performance than SLMs ($\le $ 1B). However, in general, we can only obtain the output of close-source LLMs. Based on this situation, black-box distillation is proposed to distill abilities from LLMs to SLMs. Specifically, black-box distillation first prompts LLMs to generate a distillation dataset, and then this dataset is used to fine-tune SLMs to improve their reasoning performance. For example,  
~\citet{ho-etal-2023-large} prompt LLMs to generate a CoT reasoning dataset, which was then used to fine-tune an SLM, thereby enhancing its reasoning ability. ~\citet{hsieh-etal-2023-distilling} introduce a multi-task distillation framework that enables the SLM to focus not only on predicting the final output labels but also on predicting intermediate reasoning steps. ~\citet{shridhar-etal-2023-distilling} utilize a LLM to train a problem decomposer and subproblem solver, breaking problems into subproblems and solving them individually to get the final answer. ~\citet{FuPOSK23} analyze CoT distillation, finding a trade-off: enhancing specific SLM abilities decreases their general abilities. ~\citet{abs-2305-13888} create a PoT reasoning dataset, allowing SLMs to use Python interpreters for better mathematical reasoning and error avoidance. ~\citet{zhu2024distillingmathematicalreasoningcapabilities} further formalize the reasoning process into equations and find that a diverse range of reasoning formats can effectively enhance the mathematical reasoning performance of SLMs.  In contrast to the methods mentioned above, our work presents a novel distillation approach in which one SLM  is employed to extract the core question and problem-solving information from the original question. These key points are then used to guide a second SLM in effectively solving the original question.

\section{Method}

\begin{figure*}[!ht]
	\centering
	\includegraphics[width=\textwidth]{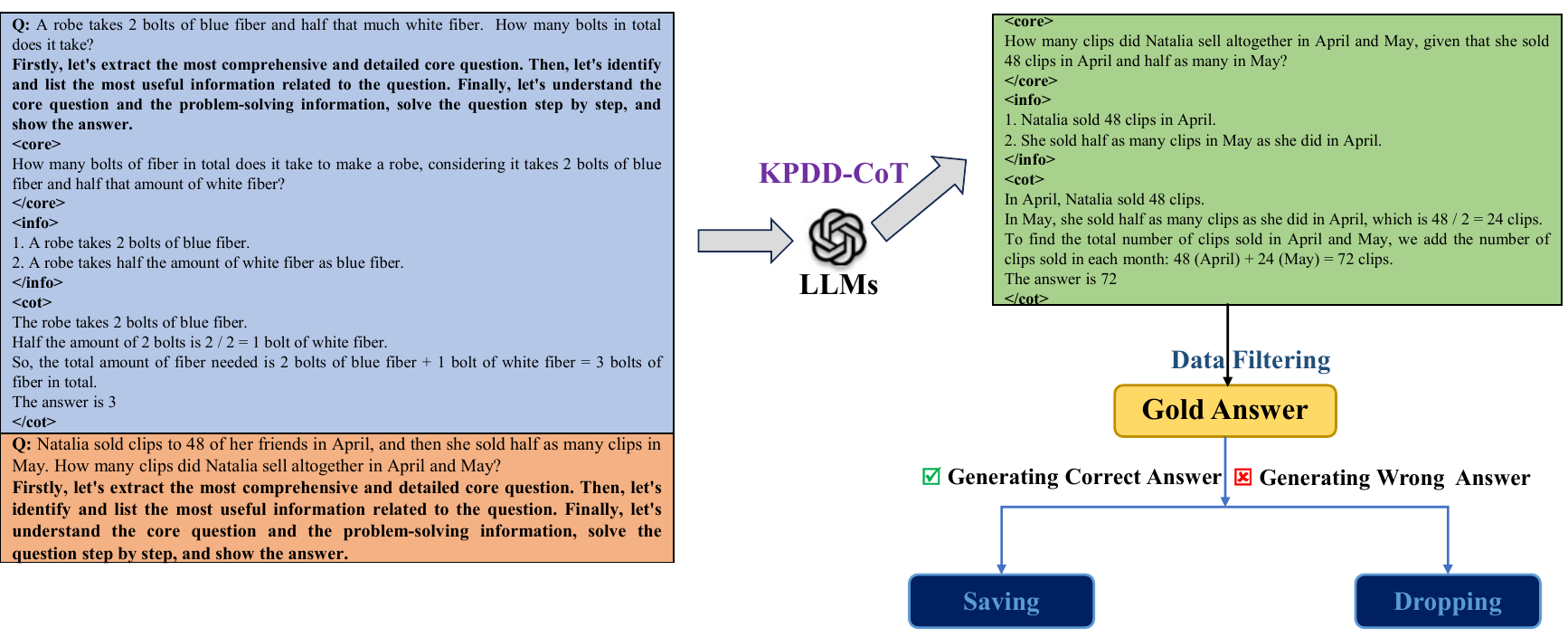}
	\caption{\textbf{Detailed data generation for KPDD-CoT.} Initially, we use few-shot prompting to guide LLMs in producing reasoning processes. We then  discard any reasoning process that does not align with the correct answers. In this way, we have constructed a high-quality KPDD-CoT reasoning distillation dataset.}
	\label{fig:cot_data_generation}
\end{figure*}

In this work, we introduce a novel distillation method for mathematical reasoning tasks called Key-Point-Driven Distillation (KPDD), structured into two stages: (1) Stage 1: KPDD distills the first SLM to extract the core question and the problem-solving information from the original question. 
(2) Stage 2: KPDD distills the second SLM to solve the original problem using the core question and problem-solving information. 

\subsection{Data Generation from LLMs}
When given a mathematical dataset, our primary task is to utilize an LLM to generate a reasoning process for each mathematical problem in the dataset. By doing so, we augment the mathematical dataset to construct the distillation dataset. Furthermore, in stage 2, our KPDD method employs two distillation approaches: one distills the SLM to generate CoT rationales for problem-solving, and the other distills the SLM to generate PoT rationales for problem-solving. In other words, our KPDD method can be divided into two approaches: KPDD-CoT and KPDD-PoT. Here is a data generation process for KPDD-CoT. Furthermore, the data generation process for KPDD-PoT is similar to that of KPDD-CoT, with the detailed procedure for KPDD-PoT provided in Appendix~\ref{sec:dgkp}.

Our method uses few-shot prompting to prompt the LLM to synthesize the reasoning dataset. Figure~\ref{fig:cot_data_generation} reveals the distillation dataset generation process. 
Specifically, we first randomly sample $k$ data pairs $(x, y)$ from a mathematical dataset $\mathcal{D}$, where $x$ represents the question and $y$ represents the answer. Then, for these sampled data, we manually construct the reasoning process $c$. Each reasoning process $c$ includes a core question, problem-solving information, and rationales in CoT format. Finally, we obtain a demonstration dataset $\mathcal{D}_c$. At the same time, we further introduce an instruction \textbf{"Firstly, let's extract the most comprehensive and detailed core question. Then, let's identify and list the most useful information related to the question. Finally, let's understand the core question and the problem-solving information, solve the question step by step, and show the answer."}. By leveraging the demonstration set and the instruction, we prompt the LLM to generate the reasoning process for the mathematical questions. The KPDD-CoT dataset generation is formalized as:
\begin{equation}
	c_i = f_\mathcal{M}(x_i, \mathcal{D}_c),
\end{equation}
where $\mathcal{M}$ denotes the LLM, $f$ represents the decoding function, and $i$ is the index in $\mathcal{D}$.

\textbf{Data Filtering}---To further improve the quality of the dataset, we attempt to filter out incorrect reasoning processes from the KPDD-CoT dataset. Specifically, for each data point in the KPDD-CoT dataset, we extract the answer from its reasoning process. If this answer matches the gold answer, we retain the data point; otherwise, we discard it as the reasoning process is deemed incorrect. By ensuring that the KPDD-CoT dataset contains only correct reasoning processes, we enhance the LLM's mathematical reasoning performance.

\subsection{Fine-tuning SLMs}
After constructing these reasoning datasets, we use them to fine-tune the SLMs. In the KPDD, we fine-tune two SLMs: the first SLM, called KPDD-CoT/PoT-Key, is used to extract the key points from the original problem, the second SLM, called KPDD-CoT/PoT-Solve, solve the original question based on these key points. Here is a fine-tuning process for KPDD-CoT.  The fine-tuning process for KPDD-PoT is similar to that of KPDD-CoT, with the detailed procedure for KPDD-PoT provided in Appendix~\ref{sec:ftkp}.

Firstly, we construct a key points subset from the KPDD-CoT dataset, denoted as $\mathcal{D}_{KP}$. Each sample in this subset can be represented as $(x, kp)$, where $x$ represents the original question and $kp$ represents key points, which consist of the core question and problem-solving information. For each training instance $(x, kp)$ from $\mathcal{D}_{KP}$, we prepend the prompt $p_{kp}$ \textbf{"Let's extract the core question and the most useful information from the question."} to the question $x$. This guides the KPDD-CoT-Key in fine-tuning to effectively extract the corresponding key points $kp$. The fine-tuning loss function can be represented as follows:
\begin{equation}
	\mathcal{L} = - \sum_{i=1}^{N} \sum_{t=1}^{T} \log P({kp}^i_t \mid {kp}^i_{< t}, x^i, p_{kp}),
\end{equation}
where $N$ is the number of examples in $\mathcal{D}_{KP}$, $p_{kp}$ is the prompt, and ${kp}_{:T}$ is the sequence of the key points. 

Then, we construct a problem-solving subset from the KPDD-CoT dataset, denoted as $\mathcal{D}_{CS}$. Each sample in this subset can be represented as a tuple $(x, kp, cs)$, where $x$ represents the original question, $kp$ represents the key points, encompassing the core question and problem-solving information, and $cs$ represents the rationales in CoT format. For each data from $\mathcal{D}_{CS}$, we integrate the original question, the key points and the prompt $p_{cs}$ \textbf{"Let's understand the core question and the problem-solving information, solve the question step by step, and show the answer."} to construct a new input. This guides the KPDD-CoT-Solve in fine-tuning to generate rationales $cs$ for solving the origin question in CoT format. The fine-tuning loss function can be represented as follows:
\begin{equation}
	\mathcal{L} = - \sum_{i=1}^{N} \sum_{t=1}^{T} \log P({cs}^i_t \mid {cs}^i_{< t}, x^i, kp^i, p_{cs}),
\end{equation}
where $N$ is the number of examples in $\mathcal{D}_{CS}$, $p_{cs}$ is the prompt, and ${cs}_{:T}$ is the sequence of the rationale in CoT format.

\subsection{Inference-time Predictions}
After fine-tuning, the process for solving a question involves two steps.  Firstly, we use the KPDD-CoT/PoT-Key SLM to extract the core question and the relevant problem-solving information from the original question. This SLM identifies and outlines the key points necessary for KPDD-CoT/PoT-Solve, enhancing the understanding of the original question. Then, based on the original question, the key points, the KPDD-CoT/PoT-Solve SLM generates rationales in either CoT or PoT format to solve the original question. For KPDD-PoT, this requires an additional Python interpreter to execute the generated Python program and obtain the final answer. The approach ensures that each SLM focuses on a specific aspect of the problem-solving process, leading to more accurate and reliable solutions.

\section{Experiments}
\begin{table}[!ht]
	\centering
    \scriptsize
	\begin{tabular}{@{}c|cc@{}}
		\toprule
		& \textbf{Dataset} & \textbf{Size} \\ \midrule
		\multirow{2}{*}{Train} & GSM8K & 7473 \\
		& (+) augmented & 29892 \\ \midrule
		\multirow{4}{*}{Test} & GSM8K & 1319 \\
		& ASDiv & 2096 \\
		& SVAMP & 1000 \\
		& MultiArith & 600 \\ \bottomrule
	\end{tabular}
 \caption{\textbf{Statistics of the datasets used in our experiments.} Augmented refers that we run 4 times data synthesis on the training set of GSM8K.}
	\label{tab:statistics_of_dataset}
\end{table}

\subsection{Dataset}
In our paper, we generate KPDD distillation datasets based on the GSM8K training set, which comprises diverse grade school math word problems~\cite{abs-2110-14168}. Then, we evaluate the mathematical reasoning performance of SLM on the GSM8K test set. Furthermore, to assess the transferability of SLMs' mathematical reasoning capabilities, we evaluate SLM on several additional mathematical datasets. These datasets include ASDiv, which contains diverse math word problems~\cite{miao-etal-2020-diverse}, SVAMP, which features math word problems with varying structures~\cite{patel-etal-2021-nlp}, and MultiArith, which consists of arithmetic word problems~\cite{roy-roth-2015-solving}. The statistics of these datasets are summarized in Table~\ref{tab:statistics_of_dataset}. This comprehensive evaluation approach ensures that the SLMs' mathematical reasoning capabilities are thoroughly tested across  a diverse range of both in-domain and out-of-domain questions, providing a robust assessment of their performance.

\subsection{Implementation}
In our paper, we use ChatGPT as our teacher LLM to generate reasoning processes for questions. Specifically, we manually construct a demonstration set containing eight demonstrations. We then use this demonstration set to prompt ChatGPT to generate four reasoning paths for each question. Each reasoning path is subsequently filtered, resulting in the creation of a KPDD dataset. Next, we use FlanT5 models—Small (60M), Base (250M), and Large (760M)~\cite{abs-2210-11416}—as our student LMs. By fine-tuning the FlanT5 models with the KPDD dataset, we aim to enhance their mathematical reasoning abilities. During the fine-tuning process, we set the learning rate to 5e-4, the batch size to 32, and the total number of training epochs to 10. The input and output lengths for KPDD-CoT/PoT-Key are set to 256 and 256, respectively, while the input and output lengths for KPDD-CoT/PoT-Solve are set to 512 and 256, respectively.

\begin{table*}[!ht]
	\centering
    \scriptsize
	\begin{tabular}{c|c|cccc|c}
		\toprule 
		\textbf{Models} & \textbf{\#Params} & \textbf{GSM8K} & \textbf{ASDiv} & \textbf{SVAMP} & \textbf{MultiArith} & \textbf{AVG}\\
		\midrule 
		\rowcolor[rgb]{0.93,0.93,0.93}
		\multicolumn{7}{l}{\textit{Proprietary Large Language Models}} \\
		GPT-4~\cite{abs-2303-08774} & - & 92.0 & 91.3 & 93.1 & - & 92.13\\
		ChatGPT & -  & 80.8 & 87.3 & 83.0 & - & 83.7\\
		Claude-2~\cite{Claude-2} & - & 85.2 & - & - & - & 85.2\\
		PaLM-2~\cite{abs-2305-10403} & 540B & 80.7 & - & - & - & 80.7\\
		\midrule 
		\rowcolor[rgb]{0.93,0.93,0.93}
		\multicolumn{7}{l}{\textit{Open-Source Large Language Models}} \\
		Llama-2~\cite{abs-2307-09288} & 7B & 13.3 & 50.7 & 38.0 & - & 34\\
		CodeLLaMA~\cite{abs-2308-12950} & 7B & 34.0 & 61.4 & 59.0 & - & 51.46\\
		Platypus-2~\cite{abs-2308-07317} & 7B & 14.4 & 47.9 & 36.7 & - & 33\\
		WizardMath~\cite{abs-2308-09583} & 7B & 54.9 & 59.1 & 57.3 & - & 57.1\\
		TORA~\cite{abs-2309-17452} & 7B & 68.8 & 73.9 & 68.2 & - & 70.3 \\
		\midrule 
		\rowcolor[rgb]{0.93,0.93,0.93}
		\multicolumn{7}{l}{\textit{Fine-tuned Small Language Models}} \\
		Ho et al.~\cite{ho-etal-2023-large} & 0.3B & 3.11 & - & - & - & 3.11\\
		Fu et al.~\cite{FuPOSK23} & 0.76B & 20.2 & 23.8 & 20.4 & 38.5 & 25.72\\
		Shridhar et al.~\cite{shridhar-etal-2023-distilling} & 0.77B & 17.89 & - & 18.14 & - & 18.01\\
		Zhu et al.~\cite{abs-2305-13888} & 0.77B & 39.2 & 51.2 & 48.2 & 79.2 & 54.45\\
		Zhu et al.~\cite{zhu2024distillingmathematicalreasoningcapabilities} & 0.77B & 42.45 & 52.81 & 49.59 & 85.5 & 57.58\\
		\midrule 
		\rowcolor[rgb]{0.93,0.93,0.93}
		\multicolumn{7}{l}{\textit{Our fine-tuned Small Language Models}} \\
		FlanT5-Small & 0.06B & 2.1 & 2.8 & 2.1 & 4.0 & 2.75\\
		(+) KPDD-CoT & & \textbf{7.73} & \textbf{13.21} & \textbf{9.8} & \textbf{9.66} & \textbf{10.1}\\
		(+) KPDD-PoT & & \textbf{21.6} & \textbf{42.27} & \textbf{34.59} & \textbf{49.83} & \textbf{37.07}\\
		\hline
		FlanT5-Base & 0.25B & 3.0 & 4.2 & 3.8 & 7.0 & 4.5\\
		(+) KPDD-CoT & & \textbf{14.32} &\textbf{19.32}& \textbf{13.9} & \textbf{23.83} & \textbf{17.84}\\
		(+) KPDD-PoT & & \textbf{36.31} & \textbf{53.24} & \textbf{47.69} & \textbf{78.33} & \textbf{53.89}\\
		\hline
		FlanT5-Large & 0.76B & 6.9 & 10.1 & 6.8 & 13.0 & 9.2\\
		(+) KPDD-CoT & & \textbf{20.92} & \textbf{27.05} & \textbf{21.5} & \textbf{40} & \textbf{27.36}\\
		(+) KPDD-PoT & & \textbf{49.05} & \textbf{60.16}& \textbf{58.09} & \textbf{90.33} & \textbf{64.4}\\
		\bottomrule 
	\end{tabular}
    \caption{\textbf{Overall Test Sets Performance.} }
	\label{tab:main_results}
\end{table*}

\subsection{Main Results}
Table~\ref{tab:main_results} showcases our method's performance on four mathematical datasets, revealing key insights:
\begin{enumerate}
	\item \textbf{KPDD-CoT Enhances Mathematical Reasoning:} 
	When FlanT5-small is used as the SLM, KPDD-CoT achieves an average accuracy improvement of 7.35\% across several mathematical reasoning tasks. With FlanT5-base as the SLM, KPDD-CoT yields an average accuracy improvement of 13.34\%. For FlanT5-large, KPDD-CoT results in an average accuracy improvement of 18.16\%. The experimental result demonstrates that KPDD-CoT can significantly enhance the mathematical reasoning performance of SLMs. We attribute this experimental result to the reason that baselines often encounters semantic misunderstanding errors that hinder the improvement of SLMs' mathematical reasoning abilities. In contrast, KPDD-CoT employs extra SLMs to extract key points (including the core question and problem-solving information) of the question and uses these key points to guide the SLMs' reasoning. This approach significantly reduces the semantic misunderstanding errors of CoTD, making KPDD-CoT better suited for improving the mathematical reasoning ability of SLMs.
	
	\item \textbf{KPDD-PoT Outperforms State-of-the-Art:} 
 When using FlanT5-small as the SLM, KPDD-PoT improves average accuracy by 34.32\% in mathematical reasoning tasks. With FlanT5-base, it increases by 49.39\%, and with FlanT5-large, by 55.2\%. This shows that KPDD-PoT helps SLMs achieve state-of-the-art reasoning accuracy. It also outperforms KPDD-CoT, demonstrating the benefits of PoT-format rationales in improving reasoning. This is because that CoTD's performance is limited not only by semantic misunderstanding errors but also by calculation errors. PoTD turns CoT rationales into Python code and uses a Python interpreter for final answers, avoiding calculation errors. KPDD-PoT also enhances question understanding, improving overall reasoning performance.
 
	
	\item \textbf{Strong Transferability of KPDD:} KPDD exhibits strong transferability. The distillation dataset of KPDD is constructed based on the GSM8K training dataset, and we evaluate our SLMs on several mathematical reasoning datasets, including the GSM8K test dataset, ASDiv dataset, SVAMP dataset, and MultiArith dataset. Our experimental results show that KPDD not only achieves good reasoning performance on the GSM8K test dataset but also performs well on the ASDiv, SVAMP, and MultiArith datasets. These results demonstrate that KPDD has strong transferability and further corroborate that SLMs do not improve their reasoning performance through data leakage.
\end{enumerate}

\begin{table*}[]
\centering
\scriptsize
\begin{tabular}{@{}cccccccccc@{}}
\toprule
\textbf{Reasoning Format} & \textbf{Category} & \textbf{Core} & \textbf{Info} & \textbf{Solve} & \textbf{GSM8K} & \textbf{ASDiv} & \textbf{SVAMP} & \textbf{MultiArith} & \textbf{AVG}\\ \midrule
\multicolumn{1}{c|}{\multirow{5}{*}{CoT}} & 1 & $\times$ & $\times$ & $\times$ & 3.0 & 4.2 & 3.8 & 7.0 & 4.5  \\
\multicolumn{1}{c|}{} &  2 & $\times$ & $\times$ & $\checkmark$ & 7.05 & 10.54 & 7.3 & 11.5 & 9.09\\
\multicolumn{1}{c|}{} & 3 & $\checkmark$ & $\times$ & $\checkmark$ & 7.35 & 12.4 & 8.2 & 11.16 & 9.77 \\
\multicolumn{1}{c|}{} & 4 & $\times$ & $\checkmark$ & $\checkmark$ & 7.73 & 12.78 & 7.7 & 11.83 & 10.01\\
\multicolumn{1}{c|}{} & 5 & $\checkmark$ & $\checkmark$ & $\checkmark$ & 8.64 & 12.92 & 8.7 & 13.83 & 11.02  \\ \midrule
\multicolumn{1}{c|}{\multirow{5}{*}{PoT}} & 1 & $\times$ & $\times$ & $\times$ & 3.0 & 4.2 & 3.8 & 7.0 & 4.5   \\
\multicolumn{1}{c|}{} & 2 & $\times$ & $\times$ & $\checkmark$ & 20.09 & 42.84 & 36.7 & 44.83 & 36.11  \\
\multicolumn{1}{c|}{} & 3 & $\checkmark$ & $\times$ & $\checkmark$ & 25.17 & 46.42 & 39.3 & 51.33 & 40.55 \\
\multicolumn{1}{c|}{} & 4 & $\times$ & $\checkmark$ & $\checkmark$ & 25.32 & 46.94 & 40.2 & 51.66 & 41.03 \\
\multicolumn{1}{c|}{} & 5 & $\checkmark$ & $\checkmark$ & $\checkmark$ & 26.08 & 47.32 & 41.5 & 53.83 & 42.18 \\ \bottomrule
\end{tabular}
\caption{\textbf{Effect of Different Components in KPDD.} We consider five different categories to analyse the effect of different components in KPDD. The experiment result shows that key points in questions can deepen SLMs' understanding of the questions, and combining several key points can provide richer information, leading to further improvements in SLMs' reasoning abilities.}
\label{tab:kpdd-components}
\end{table*}

\subsection{Effect of Different Components in KPDD}
In this subsection, we delve into the impact of various components within KPDD. We have considered five distinct categories, which include: 1. Original SLMs without any fine-tuning; 2. SLMs with original CoT/PoT distillation; 3. SLMs with core distillation combined with CoT/PoT distillation; 4. SLMs with problem-solving information distillation combined with CoT/PoT distillation; 5. SLMs with KPDD. For each of the latter four categories, we have constructed corresponding distillation datasets, each containing a single reasoning path per question. Following this, we have utilized FlanT5-base as our foundation for SLMs, and we have fine-tuned these models using the aforementioned distillation datasets. To evaluate the reasoning capabilities of these SLMs, we have tested them on the GSM8K test dataset, as well as on the ASDiv, SVAMP, and MultiArith datasets.

Tables~\ref{tab:kpdd-components} present the results of our experiments, from which we make several observations: (1) We observe a significant performance improvement in Category 2 compared to original SLMs. Specifically, under CoT reasoning, Category 2 achieves an average accuracy gain of 4.59\% across multiple datasets, while under PoT reasoning, it achieves a substantial average accuracy improvement of 31.61\%. These experimental results indicate that CoTD and PoTD can markedly enhance the mathematical reasoning ability of SLMs. (2) We find that Categories 3 and 4 exhibit a further performance increase relative to Category 2. Specifically, in the context of CoT reasoning, Categories 3 and 4 achieve average accuracy gains of 0.77\% and 0.92\% respectively over Category 2 across multiple datasets. Under PoT reasoning, the gains are more pronounced with Categories 3 and 4 achieving average accuracy improvements of 4.44\% and 4.92\% respectively. This suggests that SLMs can deepen their understanding of questions by focusing on key points, thereby further enhancing their mathematical reasoning ability. (3) In Category 5, we combine the core questions with the problem-solving information to guide SLMs in addressing the questions. The results are promising: Category 5 achieves an average accuracy of 11.02\% under CoT reasoning and a remarkable 42.18\% under PoT reasoning across multiple datasets. This indicates that key points in questions play a crucial role in boosting the reasoning capabilities of SLMs, and that combining several key points provides richer information, leading to further improvements in their reasoning abilities.

\begin{figure}[!ht]
	\centering
	\includegraphics[width=\columnwidth]{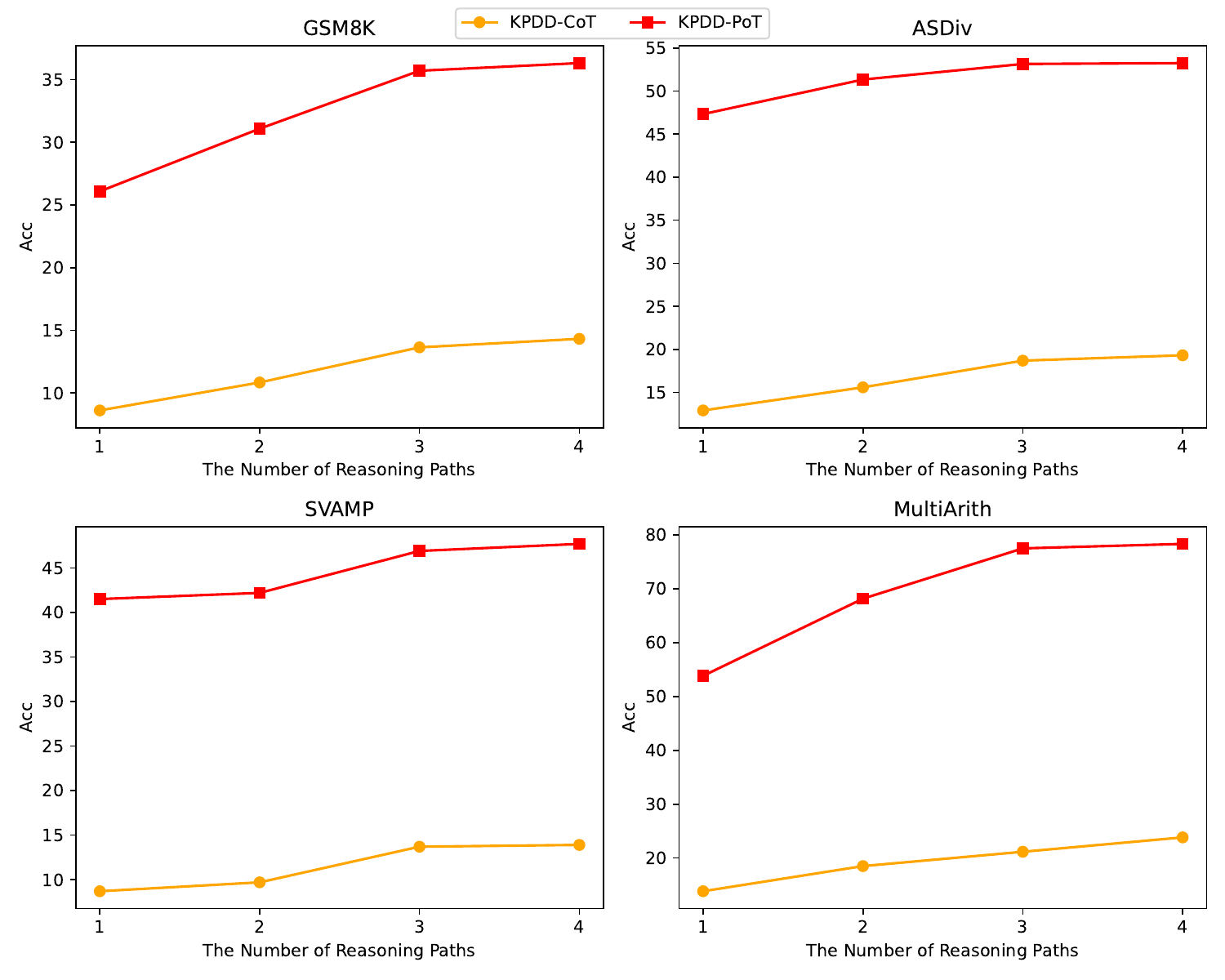}
	\caption{\textbf{Effect of Reasoning Paths.} We fine-tune FlanT5-Base with different reasoning paths to analyse the effect of reasoning paths. The experiment results shows that diverse reasoning paths can improve SLMs' reasoning performance.}
	\label{fig:reasoning_path_effect}
\end{figure}

\subsection{Diverse Reasoning Paths Improve SLMs' Reasoning Performance}
In this subsection, we primarily explore the impact of the diversity of reasoning paths on the mathematical reasoning performance of SLMs. Specifically, we fine-tune FlanT5-base using KPDD datasets with varying numbers of reasoning paths and then evaluate the fine-tuned SLMs on several mathematical reasoning datasets. By analyzing their mathematical reasoning performance, we assess how the diversity of reasoning paths affects the SLMs' capabilities in mathematical reasoning.

Figure~\ref{fig:reasoning_path_effect} reveals the relevant experimental results. As the number of reasoning paths in the distillation dataset increases, the SLM tends to achieve better performance. For example, when fine-tuning FlanT5-base with the KPDD-CoT dataset containing a single reasoning path, it achieved an accuracy of 8.64\% on the GSM8K test dataset, 12.92\% on ASDiv, 8.7\% on SVAMP, and 13.83\% on MultiArith. However, when fine-tuning FlanT5-base with the KPDD-CoT dataset containing four reasoning paths, it achieved accuracies of 14.32\% on the GSM8K test dataset, 19.32\% on ASDiv, 13.9\% on SVAMP, and 23.83\% on MultiArith. Similarly, when fine-tuning SLMs with KPDD-PoT datasets containing different reasoning paths, the experimental results were consistent with those of KPDD-CoT. These results indicate that a mathematical reasoning dataset with diverse reasoning paths can effectively enhance the mathematical reasoning performance of SLMs.

\begin{figure*}[!ht]
	\centering
	\includegraphics[width=\textwidth]{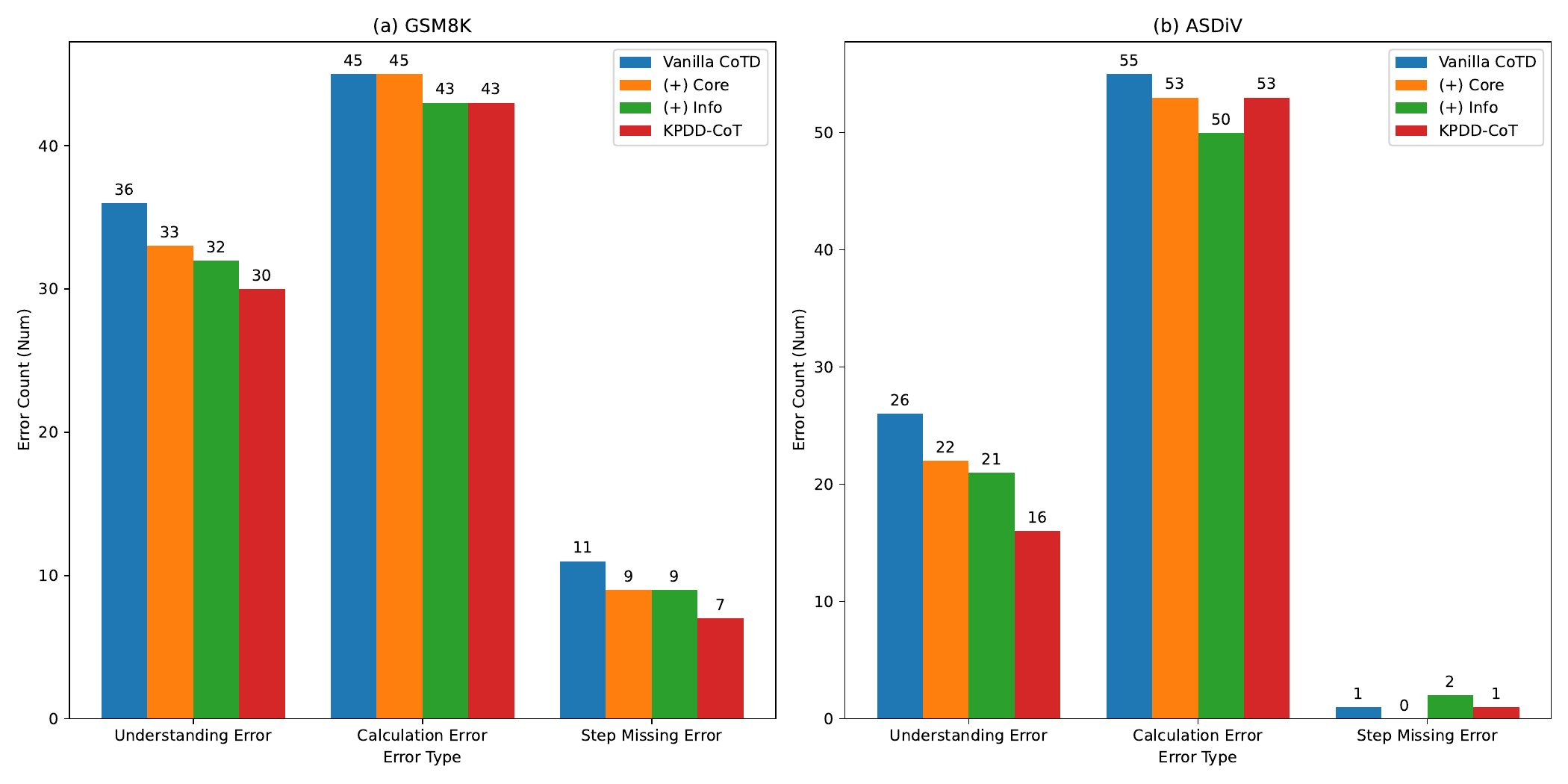}
	\caption{\textbf{Error Analysis for SLMs.} We conducted an error analysis of four different categories of distillation methods. The experiment results show that integrating multiple key points of the questions can significantly reduce SLMs' understanding errors, enhance the comprehension of the questions and further improve the reasoning performance of SLMs.}
	\label{fig:error_analysis}
\end{figure*}

\subsection{Error Analysis}
In this subsection, our aim is to verify whether KPDD can indeed reduce semantic misunderstanding errors. KPDD-PoT implicitly includes the reasoning process within its rationales, making it challenging to conduct error analysis on rationales in PoT format. Conversely, rationales in CoT format explicitly contain the reasoning steps, allowing us to clearly understand how the SLM solves the questions step by step, thus facilitating error analysis. Therefore, in this part, we focus on error analysis for rationales in CoT format. To achieve our goal, we randomly sample 100 examples from GSM8K/ASDiV and perform error analysis on the questions with incorrect answers. For a better understanding of KPDD's effect, we also consider three other scenarios: (1) vanilla CoTD, (2) reasoning that combines CoTD and core question extraction, and (3) reasoning that combines CoTD and problem-solving information extraction. Furthermore, to simplify our analysis, we use FlanT5-base as our SLMs, and the corresponding reasoning datasets still contain a single reasoning path per question.

The detailed quantitative results are illustrated in Figure~\ref{fig:error_analysis}. By analyzing the experimental results, we found that: (1) \textbf{Calculation Errors and Understanding Errors are The Major Errors:} The experiment results indicate that calculation errors and misunderstanding errors are the primary factors limiting CoTD's mathematical reasoning performance.
Specifically, vanilla CoTD on the GSM8K dataset showed 36 understanding errors, 45 calculation errors, and 11 step missing errors. Similar results were observed in the ASDiV dataset. This explains why PoTD achieves better reasoning performance than CoTD: PoTD converts vanilla rationales into Python programs, delegating the calculation process to an external Python interpreter to avoid calculation errors. (2) \textbf{Reduction of Understanding Errors with Key Points:} Introducing key points of the original questions effectively reduces understanding errors. Specifically, when core questions were introduced in vanilla CoTD, the number of understanding errors on the GSM8K dataset decreased to 33, and on the ASDiV dataset, it decreased to 22. When problem-solving information was introduced in vanilla CoTD, the number of understanding errors decreased to 32 on GSM8K and to 21 on ASDiV. These results indicate that key points of the original questions help SLMs better understand the questions, thereby reducing understanding errors and improving reasoning performance. (3) \textbf{Further Reduction of Understanding Errors with Multiple Key Points:} Combining multiple key points can further reduce understanding errors. Specifically, KPDD reduced the number of understanding errors to 30 on GSM8K and to 16 on ASDiV. This suggests that KPDD's method of integrating multiple key points can deepen SLMs' understanding of the original questions, further reducing understanding errors and enhancing reasoning performance.

\section{Conclusion}
In this paper, we propose Key-Point-Driven Distillation (KPDD) for enhancing mathematical reasoning in Small Language Models (SLMs). Our approach leverages the extraction of key points from questions to improve understanding and reduce errors in reasoning tasks. Experimental results demonstrate that KPDD significantly reduces understanding errors compared to conventional mathematical reasoning distillation method. 
However, PoTD inherently integrates the reasoning process into the generated program, which complicates error analysis. In the future, we will investigate additional methods for error analysis to enhance the evaluation of PoTD's performance.


\bibliography{custom}

\clearpage
\appendix
\begin{figure*}[!ht]
	\centering
	\includegraphics[width=\textwidth]{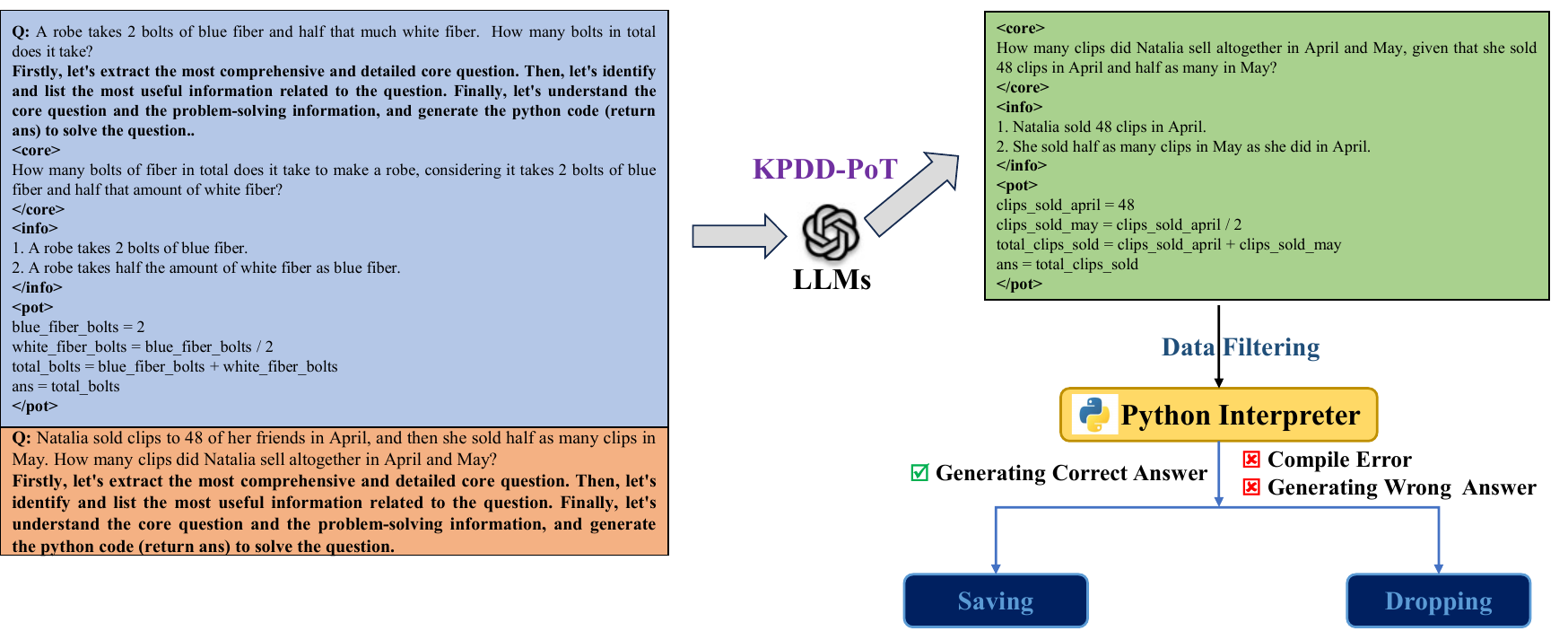}
	\caption{\textbf{Detailed data generation for KPDD-PoT.} Similar to KPDD-CoT, we prompt LLMs to generate the reasoning process in the KPDD-PoT format. We extract the rationale in the PoT format from this process and run it through a Python interpreter. If there are errors or incorrect answers, we discard the reasoning process. Finally, we constructed a high-quality KPDD-PoT reasoning dataset.}
	\label{fig:pot_data_generation}
\end{figure*}

\section{Data Generation for KPDD-PoT}
\label{sec:dgkp}
Similar to KPDD-CoT, we first sample $k$ examples from a mathematical dataset $\mathcal{D}$, then use these samples to manually create a demonstration set $\mathcal{D}_p$. Each reasoning process in the demonstration set $\mathcal{D}_p$ includes a core question, problem-solving information, and rationales in PoT format.  By utilizing the  demonstration set $\mathcal{D}_p$ and the instruction, we prompt LLMs  to generate reasoning processes in for questions. Figure~\ref{fig:pot_data_generation} shows the data synthesis process for KPDD-PoT, and the KPDD-PoT dataset generation process can be formalized as:
\begin{equation}
	p_i = f_\mathcal{M}(x_i, \mathcal{D}_p).
\end{equation}

\textbf{Data Filtering}---Similar to data filtering for KPDD-CoT, we also apply filtering to the KPDD-PoT dataset. Specifically, for each data point in the KPDD-PoT dataset, we extract the rationale in PoT format from its reasoning process. An external Python interpreter is then used to run the rationale. If the obtained answer do not match the gold answer, it indicate that the reasoning process is incorrect. Consequently, we remove the data point with the flawed reasoning process from the KPDD-PoT dataset, thereby enhancing the dataset's quality.

\section{Fine-tuning SLMs for KPDD-PoT}
\label{sec:ftkp}
In KPDD-PoT, aside from replacing the KPDD-CoT dataset with the KPDD-PoT dataset, the fine-tuning method for KPDD-PoT-Key remains consistent with that of KPDD-CoT-Key. However, the fine-tuning method of KPDD-PoT-Solve is different with KPDD-CoT-Solve. The main difference between them is the input instruction. Specifically, when fine-tuning KPDD-PoT-solve, the input instruction is \textbf{"Let's understand the core question and the problem-solving information, and generate a python program to solve the question."} This instruction guides the model to not only understand the core question and the problem-solving information but also to generate Python code that can get the final answer. This approach leverages the model's ability of code generation, which can be particularly effective for solving mathematical problems programmatically.

Moreover, the fine-tuning loss functions for the SLMs in KPDD-PoT are identical to those in KPDD-CoT. This ensures that the optimization process remains consistent across both methods, focusing on minimizing the discrepancies between the model's output and the expected solutions.

\end{document}